\documentclass{article}

\usepackage{arxiv}

\usepackage[utf8]{inputenc} 
\usepackage[T1]{fontenc}    
\usepackage{url}            
\usepackage{booktabs}       
\usepackage{amsfonts}       
\usepackage{nicefrac}       
\usepackage{microtype}      

\usepackage[onehalfspacing]{setspace}

\usepackage{graphicx}
\usepackage{subcaption}
\usepackage{color}
\usepackage{amsmath}
\usepackage{amssymb}
\usepackage{breqn}
\usepackage{verbatim}
\usepackage{balance}
\usepackage{tabularx}
\usepackage{array}
\usepackage{caption}
\newcolumntype{?}{!{\vrule width 1.5pt}}

\usepackage[ruled, linesnumbered]{algorithm2e}

\SetCommentSty{mycommfont}
\SetKwComment{Comment}{$//$\ }{}
\usepackage{float}

\graphicspath{{imgs/}}

\title{Feature Guided Search for Creative Problem Solving Through Tool Construction}

\author{
  Lakshmi Nair\\
  College of Computing\\
  Georgia Institute of Technology\\
  Atlanta, GA 30318 \\
  \texttt{lnair3@gatech.edu} \\
   \And
  Sonia Chernova\\
  College of Computing\\
  Georgia Institute of Technology\\
  Atlanta, GA 30318 \\
  \texttt{chernova@gatech.edu} \\
}

\begin{document}
\maketitle

\begin{abstract}
Robots in the real world should be able to adapt to unforeseen circumstances. Particularly in the context of tool use, robots may not have access to the tools they need for completing a task. In this paper, we focus on the problem of tool construction in the context of task planning. We seek to enable robots to construct replacements for missing tools using available objects, in order to complete the given task. We introduce the Feature Guided Search (FGS) algorithm that enables the application of existing heuristic search approaches in the context of task planning, to perform tool construction efficiently. FGS accounts for physical attributes of objects (e.g., shape, material) during the search for a valid task plan. Our results demonstrate that FGS significantly reduces the search effort over standard heuristic search approaches by $\approx 93\%$ for tool construction.
\end{abstract}


\section{Introduction}
Humans often show remarkable improvisation capabilities, particularly in times of crises. The makeshift carbon dioxide filter constructed on board the Apollo 13 \cite{cass2005apollo}, and the jury-rigged ventilators built to combat equipment shortages during COVID-19 \cite{turner2020thinking}, are examples of human ingenuity in the face of uncertainty. In addition to humans, other primates and certain species of birds have also been shown to creatively accomplish tasks by constructing tools from available objects, such as sticks and stones \cite{stout2011stone, toth1993pan, jones1973tool}. While the capability to construct tools is often regarded as a hallmark of sophisticated intelligence, similar improvisation capabilities are currently beyond the scope of existing robotic systems. The ability to improvise and construct necessary tools can greatly increase robot adaptability to unforeseen circumstances, enabling robots to handle any uncertainties or equipment failures that may arise \cite{atkeson2018happened}. 

In this paper, we focus on the problem of tool construction in the context of task planning. Specifically, we address the scenario in which a robot is provided with a task that requires certain tools that are missing or unavailable. The robot must then derive a task plan that involves constructing an appropriate replacement tool from objects that are available to it, and use the constructed tool to accomplish the task. Existing work that addresses the problem of planning in the case of missing tools focuses on directly substituting the missing tool with available objects \cite{boteanu2015towards,agostini2015using,nyga2018grounding}. In contrast, this is the first work to address the problem through the construction of replacement tools, by introducing a novel approach called \textbf{Feature Guided Search (FGS)}. FGS enables efficient application of existing heuristic search algorithms in the context of task planning in order to perform tool construction by accounting for physical attributes of objects (e.g., shape, material) during the search for a valid task plan. 

Heuristic search algorithms, such as $A^*$ and enforced hill-climbing (EHC), have been successfully applied to planning problems in conjunction with heuristics such as cost-optimal landmarks \cite{karpas2009cost} and fast-forward \cite{hoffmann2001ff} respectively. However, the application of heuristic search algorithms to perform tool construction in the context of task planning can be challenging. For example, consider a task where the goal of the robot is to hang a painting on the wall. In the absence of a hammer that is required for hammering a nail to complete the task, the robot may choose to construct a replacement for the hammer using the objects available to it. How does the robot know which objects should be combined to construct the replacement tool? One possible solution is for the user to manually encode the correct object combination in the goal definition, and the search procedure would find it. However, it is impractical for the user to know and encode the correct object combination to use, for all the objects that the robot could possibly encounter. Alternatively, the robot can autonomously attempt every possible object combination until it finds an appropriate tool construction for completing the task. However, this would require a prohibitive number of tool construction attempts. Further, what if the robot \textit{cannot} construct a good replacement for a hammer using the available objects, but can instead construct a makeshift screwdriver to tighten a screw and complete the task? In this case, the task plan would also have to be adapted to appropriately use the constructed tool, i.e., ``tighten'' a screw with the screwdriver instead of ``hammering'' the nail. In order to address these challenges, FGS combines existing planning heuristics with a score that is computed from input point clouds of objects indicating the best object combination to use for constructing a replacement tool. The chosen replacement tool then in turn guides the correct action(s) to be executed for completing the task (e.g., ``tighten'' vs. ``hammering''). Hence, our algorithm seeks to: a) eliminate the need for the user to specify the correct object combination, thus enabling the robot to autonomously choose the right tool construction based on the available objects and the task goal, b) minimize the number of failed tool construction attempts in finding the correct solution, and c) adapt the task plan to appropriately use the constructed replacement tool. 

Prior work by Nair et al. introduced a novel computational framework for performing tool construction, in which the approach takes an input action, e.g., ``flip'', in order to output a ranking of different object combinations for constructing a tool that can perform the specified action, e.g., constructing a spatula \cite{nair2019toolconstr,nair2019autonomous}. For performing the ranking, the approach scored object combinations based on the shape and material properties of the objects, and whether the objects could be attached appropriately to construct the desired tool. In contrast, this work focuses on the application of heuristic search algorithms such as $A^*$, to the problem of tool construction in the context of task planning. In this case, the robot is provided an input \textit{task}, e.g., ``make pancakes'', that requires tools that are inaccessible to the robot, e.g., a missing spatula. The robot must then output a \textit{task plan} for making pancakes, that involves constructing an appropriate replacement tool from available objects, and adapting the task plan to use the constructed tool for completing the task. Thus, prior work takes an action as input, and outputs a ranking of object combinations. In contrast, our work takes a task as input, and outputs a task plan that involves constructing and using an appropriate replacement tool. Hence, our work relaxes a key assumption of the prior work that requires the input action to be specified. Our approach directly uses the score computation methodology described in prior work \cite{nair2019toolconstr, nair2019autonomous}, but combines it with planning heuristics to integrate tool construction within a task planning framework. Our core contributions in this paper include:
\begin{itemize}
    \item Introducing the Feature Guided Search (FGS) approach that integrates reasoning about physical attributes of objects with existing heuristic search algorithms for efficiently performing tool construction in the context of task planning. 
    \item Improving upon prior work by enabling the robot to automatically choose the correct tool construction and the appropriate action based on the task and available objects, thus eliminating the need to explicitly specify an input action as assumed in prior work.
\end{itemize}

We evaluate our approach in comparison to standard heuristic search baselines, on the construction of six different tool types (hammer, screwdriver, ladle, spatula, rake, and squeegee), in three task domains (wood-working, cooking, and cleaning). Our results show that FGS outperforms the baselines by significantly reducing computational effort in terms of number of failed construction attempts. We also demonstrate the adaptability of the task plans generated by FGS based on the objects available in the environment, in terms of executing the correct action with the constructed tool. 

\section{Related Work}
Prior work by Sarathy and Scheutz have focused on formalizing creative problem solving in the context of planning problems \cite{sarathy2017macgyver, sarathy2018macgyver}. They define the notion of ``Macgyver-esque'' creativity as embodied agents that can ``\textit{generate, execute, and learn strategies for identifying and solving seemingly unsolvable real-world problems}'' \cite{sarathy2017macgyver}. They formalize Macgyvering problems (MGP) with respect to an agent $t$, as a planning problem in the agent's world $\mathbb{W}_t$, that has a goal state $g$ currently unreachable by the agent. As described in their work, solving an MGP requires a domain extension or contraction through perceiving the agent's environment and self. Prior work by Sarathy also provide an in-depth discussion of the cognitive processes involved in creative problem solving in detail, by leveraging existing work in Neuroscience \cite{sarathy2018real}. Prior work by Olte{\c{t}}eanu and Falomir has also looked at the problem of Object Replacement and Object Composition (OROC) situated within a cognitive framework called, the Creative Cognitive Framework (CreaCogs) \cite{oltecteanu2016object}. Their work utilizes a knowledge base that semantically encodes object properties and relationships in order to reason about alternative uses for objects to creatively solve problems. The semantic relationships themselves are currently encoded a-priori. Similar work by Freedman et al. has focused on the integration of analogical reasoning and automated planning for creative problem solving by leveraging semantic relationships between objects \cite{freedman2020}. They present the Creative Problem Solver (CPS), that uses large-scale knowledge bases to reason about alternate uses of objects for creative problem solving. In contrast to reasoning about objects, prior work by Gizzi et al. has looked at the problem of discovering new actions for creative problem solving, enabling the robot to identify previously unknown actions \cite{gizzi2019creative}. Their work applies action segmentation and change-point detection to previously known actions to enable a robot to discover new actions. The authors then apply breadth-first search and depth-first search in order to derive planning solutions using the newly discovered actions. 

In related work, Erdogan and Stilman described techniques for Automated Design of Functional Structures (ADFS), involving construction of navigational structures, e.g., stairs or bridges. They introduce a framework for effectively partitioning the solution space by inducing constraints on the design of the structures. Further, Tosun et al. have looked at planning for construction of functional structures by modular robots, focusing on identifying features that enable environment modification in order to make the terrain traversable \cite{tosun2018perception}. In similar work, Saboia et al. have looked at modification of unstructured environments using objects, to create ramps that enhance navigability \cite{saboia2018autonomous}. More recently, Choi et al. extended the cognitive architecture ICARUS to support the creation and use of functional structures such as ramps, in abstract planning scenarios \cite{choi2018creating}. Their work focuses on using physical attributes of objects that is encoded a-priori, such as weight and size, in order to reason about the construction and stability of navigational structures. More broadly, these approaches are primarily focused on improving robot navigation through environment modification as opposed to construction of tools. Some existing research has also explored the construction of simple machines such as levers, using environmental objects \cite{stilman2014robots, levihn2014using}. Their work formulates the construction of simple machines as a constraint satisfaction problem where the constraints represent the relationships between the design components. The constraints in their work limit the variability of the simple machines that can be constructed, focusing only on the placement of components relative to one another, e.g., placing a plank over a stone to create a lever. Additionally, Wicaksono and Sheh have focused on using 3D printing to fabricate tools from polymers \cite{wicaksono17towards}. Their work encodes the geometries of specific sub-parts of tools, and enables the robot to experiment with different configurations of the fabricated tools to evaluate their success for accomplishing a task. 

The work described in this paper differs from the research described above in that we focus on creative problem solving through tool construction. Specifically, we focus on planning tasks in which the required tools need to be constructed from available objects. Two key aspects of our work that further distinguish it from existing research include: i) sensing and reasoning about physical features of objects, such as shape, material, and the different ways in which objects can be attached, and ii) improving the performance of heuristic search algorithms for tool construction in the context of task planning, by incorporating the physical properties of objects during the search for a task plan.

\section{Approach}
\label{sec:approach}

In this section, we begin by discussing some background details regarding heuristic search, followed by specific implementation details of FGS.

\subsection{Heuristic Search}
Heuristic search algorithms are guided by a cost function $f(s) = g(s) + h(s)$, where $g(s)$ is the best-known distance from the initial state to the state $s$, and $h(s)$ is a heuristic function that estimates the cost from $s$ to the goal state. An admissible heuristic never overestimates the path cost from any state $s$ to the goal \cite{hart1968formal, zhang2009search}. A consistent heuristic holds the additional property that, if there is a path from a state $x$ to a state $y$, then $h(x) \leq d(x,y) + h(y)$, where $d(x,y)$ is the distance from x to y \cite{hart1968formal}. Most heuristic search algorithms, including $A^*$, operate by maintaining a priority queue of states to be expanded (the open list), sorted based on the cost function. At each step, the state with the least cost is chosen, expanded, and the successors are added to the open list. If a successor state is already visited, the search algorithm may choose to re-expand the state, \textit{only if} the new path cost to the state is lesser than the previously found path cost \cite{bagchi1983search}. The search continues until the goal state is found, or the open list becomes empty, in which case no plan is returned.

\subsection{Feature Guided Search}

\begin{algorithm}[!t]
\setstretch{1.35}
    \SetKwInOut{Input}{input}\SetKwInOut{Output}{output}
    \SetKw{Continue}{continue}
    \SetCommentSty{itshape}
	\DontPrintSemicolon
    \SetKwFunction{FMain}{Search}
    \SetKwProg{Fn}{Function}{:}{}
    \Fn{\FMain{$\mathcal{P_D}, \mathcal{P_T}$, trust=true}}{

        $s_i, s_g =$ extractStates$(\mathcal{P_T})$
        
        $A =$ extractActions$(\mathcal{P_D})$
        
        $O =$ extractObjects$()$
        
        $O_{reject} = [$ $]$
        
        openList $= [$ $]$
        
        setPathCost($s_i, 0$)        \Comment*[r]{Set initial state's g(s) and f(s) to 0}
        
        openList.\texttt{add}($s_i, 0$)
        
        
        \While{OpenList \textbf{not} empty}{
            currState = $argmin_s(f(s)) \ \forall \ s \in$ openList
            
            openList.\texttt{pop}(currState)
            
            \uIf{currState = $s_g$}{
                \Return extractPlan($s_g$, $s_i$)
            }
            
            nextStates = getNext(currState, A)
            
            \For{$(s, a, O_a) \in$ nextStates}{
            
            $g(s) = $ computePathCost($s$, currState) \Comment*[r]{Get current path cost}
                
                $c(s) = $ getPathCost($s$) \Comment*[r]{Get previous best known path cost}
                
                \uIf{$g(s) \geq c(s)$}{
                    \Continue
                }
                \Else{
                    setPathCost($s, g(s)$) \Comment*[r]{Update lower costs as new paths are found}
                }

                $h(s) = $ computeHeuristic($s$)
                
                $\phi(s) = $ \textbf{featureScore}($s, a, O_a, trust$) \Comment*[r]{Compute the feature score: Algorithm 2}
                
                \uIf{$\phi(s) = -\infty$}{
                    $O_{reject}$.\texttt{add}$(O_a, a)$ \Comment*[r]{Track rejected combinations}
                }
                
                $f(s) = g(s) + h(s) - \phi(s)$
                
                \uIf{$f(s) = \infty$}{
                    \Continue
                }
                
                openList.\texttt{add}($s, f(s)$)
            }
        }
        
        \uIf{$O_{reject}$ not $\emptyset$}{
        
            \FMain{$\mathcal{P_D}, \mathcal{P_T}, trust=false$} \Comment*[r]{Re-attempt without trusting all sensors}
        }
        
        \Return $\emptyset$ \Comment*[r]{No plan found}
    
    }
	\caption{Feature Guided $A^*$ Search}
	\label{alg:1}
\end{algorithm}

We now describe the implementation of FGS\footnote{All source code including problem and domain definitions, are publicly available at \url{https://github.com/Lnair1993/Tool_Macgyvering}}. For the purposes of this explanation, we present our work in the context of $A^*$, though our approach can be easily extended to other heuristic search algorithms as demonstrated in our experiments. Let $S$ denote the set of states, $A$ denote the set of actions, $\gamma$ denote state transitions, $s_i$ denote the initial state, and $s_g$ denote the goal state. For the planning task, we consider the problem to be specified in Planning Domain Definition Language (PDDL) \cite{mcdermott1998pddl}, consisting of a domain definition $\mathcal{P_D} = (S, A, \gamma)$, and a problem/task definition $\mathcal{P_T} = (\mathcal{P_D}, s_i, s_g)$. Further, we use $O$ to denote a set of $n$ objects in the environment available for tool construction, $O = \{o_1, o_2, ... o_n\}$. 

Since our work focuses on tools, we assume that some action(s) in $A$ are parameterized by a set of object(s) $O_a \subseteq O$, that are used to perform the action. Specifically for tool construction, we explicitly define an action ``join$(O_a)$'', where $O_a = \{o_1, o_2, ... o_m\}, m \leq n$, parameterized by objects that can be joined to construct a tool for completing the task. For example, the action ``join-hammer$(O_a)$'' allows the robot to construct a hammer using the objects $O_a$ that parameterize the action. For actions that are not parameterized by any object, $O_a = \emptyset$. Our approach seeks to assign a ``feature score'' to the objects in $O_a$, indicating their fitness for performing the action $a$. Thus, given different sets of objects $O_a$ that are valid parameterizations of $a$, the feature score can help guide the search to generate task plans that involve using the objects that are most appropriate for performing the action. In the context of tool construction, the feature score guides the search to generate task plans that involve joining the most appropriate objects for constructing the replacement tool, given the objects available in the environment. Feature scoring can also reject objects that are unfit for tool construction.

Our approach is presented in Algorithm 1. The search algorithm extracts information regarding the initial state $s_i$ and goal state $s_g$ from the task definition (Line 2). The set of actions $A$ is extracted from the domain definition $\mathcal{P_D}$ (Line 3). The agent extracts the objects in its environment from an RGB-D observation of the scene through point cloud segmentation and clustering (Line 4).  We initialize the open list ($openList$) as a priority queue with the initial state $s_i$ and cost of 0 (Lines 6-8). Lines 9-32 proceed according to the standard $A^*$ search algorithm, except for the computation of the feature score in Line 24. While the open list is not empty, we select the state with the lowest cost function (Line 10,11). If the goal is found, the plan is extracted (Lines 12-13), otherwise the successor states are generated (Line 14). For each successor state $s$, the algorithm computes the path cost $g(s)$ from the current state $currState$ to $s$ (Line 16). The algorithm then retrieves the best known path cost $c(s)$ for the state from its previous encounters (Line 17). If the state was not previously seen, $c(s) = \infty$. In Lines 18-22, the algorithm compares the best known path cost to the current path cost, and updates the best known path cost if $g(s) < c(s)$. The algorithm then computes the heuristic $h(s)$ (Line 23), and the feature score $\phi(s)$ (Line 24). The algorithm also maintains a list of object combinations that were rejected by feature scoring (i.e., assigned a score of $-\infty$), in $O_{reject}$ (Line 26). The final cost is computed as $f(s) = g(s) + h(s) - \phi(s)$ (Line 27; We expand more on our choice of cost function in Section \ref{sec:final}). If $f(s) \neq \infty$, then the state is added to the open list, prioritized by the cost. The search continues until a plan is found, or exits if the open list becomes empty. If no plan was found, the search is reattempted (Line 34) by modifying the feature score computation (described in Sections \ref{sec:visual_score}). If all search attempts fail, the planner returns a failure with no plan found. In the following section we discuss the computation of the feature score in detail.

\subsection{Feature Score Computation}
\label{sec:visual_score}

In this section, we describe the computation of the feature score for a given set of objects $O_a$ that parameterize an action $a$. Note that, in this work the feature score computation focuses on the problem of tool construction. However, FGS can potentially be extended to other problems such as tool substitution, by computing similar feature scores as described in prior work \cite{abelha2016model,shrivatsav2019tool}. Given $n$ objects, tool construction presents a challenging combinatorial problem with a state space of size $^nP_m$, assuming that we wish to construct a tool with $m \leq n$ objects. Thus, $O_a = \{o_1, o_2, ... o_m\}$ denotes a specific permutation of $m$ objects. Prior work introduced a multi-objective function for evaluating the fitness of objects for tool construction \cite{nair2019autonomous,nair2020ijrr}, that we apply in this work for feature scoring. The proposed multi-objective function included three considerations: a) shape fitness of the objects for performing the action, b) material fitness of the objects for performing the action, and c) evaluating whether the objects in $O_a$ can be attached to construct the tool. 

The calculation of each of the three metrics above relies on real-world sensing, which can be noisy. This can result in false negative predictions, that eliminate potentially valid object combinations from consideration. In particular, prior work has shown that false negatives in material and attachment predictions have caused $\approx 4\%$ of tool constructions to fail \cite{nair2019autonomous}. To address the problem of false negatives in material and attachment predictions, we introduce the notion of ``sensor trust'' in this work. Prior work that has looked at accounting for sensor trust has introduced the notion of ``trust weighting'' to use continuous values to appropriately weigh the sensor inputs \cite{pierson2016adaptive}. In contrast, the sensor trust parameter in our work is a \textit{binary value} that determines whether the material and attachment predictions should be believed by the robot and included in the feature score computation. This is because material and attachment scores are hard constraints and not continuous, i.e., they are $-\infty$ for objects that are not suited for tool construction (we describe this further in later sections). Hence, a continuous weighting on the material and attachment scores is not appropriate for our work.  

Our feature score computation approach is described in Algorithm 2. For actions that are not parameterized by objects, the approach returns 0 (Lines 2-3). If the trust parameter is set to $true$, the feature score computation incorporates shape, material, and attachment predictions. (Lines 5-12 of Algorithm 2; Section \ref{sec:trust} for details). If the trust parameter is set to $false$, the feature score computation only includes shape scoring (Lines 14-19 of Algorithm 2; Section \ref{sec:notrust} for details). Thus, we describe two modes of feature score computation that is influenced by the sensor trust parameter. In the following sections, we briefly describe the computation of shape, material and attachment predictions, and for a more detailed implementation of each method, we refer the reader to \cite{nair2019autonomous} and \cite{erickson2019classification}. 

\begin{algorithm}[t]
\setstretch{1.35}
    \SetKwInOut{Input}{input}\SetKwInOut{Output}{output}
    \SetCommentSty{itshape}
	\DontPrintSemicolon
    \SetKwFunction{FMain}{FeatureScore}
    \SetKwProg{Fn}{Function}{:}{}
    \Fn{\FMain{$s, a, O_a, trust=true$}}{
        
        \uIf{$O_a$ \textbf{is} empty}{
            \Return 0
        }
        
        \uIf{trust}{
        
            \uIf{canAttach$(O_a, a)$}{

                $\phi^{s}_{shape}(O_a) =$ ShapeFit$(O_a, a)$ \Comment*[r]{Sensors are fully trusted - Section \ref{sec:trust}}
        
                $\phi^{s}_{mat}(O_a) =$ MaterialFit$(O_a, a)$ 
            
                $\phi(s) = \lambda_1*\phi^{s}_{shape} (O_a) + \lambda_2*\phi^{s}_{mat} (O_a)$ \Comment*[r]{The weighted sum is assigned to s}
            
                \Return $\phi(s)$
        
            }
        
            \Else{
        
                \Return $-\infty$
        
            }
        }
        
        \Else{
            
            \uIf{$(O_a, a) \in O_{reject}$}{
        
                $\phi^{s}_{shape}(O_a) =$ ShapeFit$(O_a, a)$ \Comment*[r]{Not fully trust sensors - Section \ref{sec:notrust}}
                
                \Return $\phi^{s}_{shape}(O_a)$ \Comment*[r]{Evaluate objects that were previously rejected}
            }
            \Else{
                \Return $-\infty$
            }
        }
    }
	\caption{Feature Score Computation}
	\label{alg:2}
\end{algorithm}


\subsubsection{Shape Scoring}
\label{sec:shape}

Shape scoring seeks to predict the shape fitness of the objects in $O_a$ for performing the action $a$. This is indicated by the $ShapeFit()$ function in Algorithm 2. In this work, we consider tools to have action parts and grasp parts\footnote{As in prior work, this covers the vast majority of tools \cite{myers2015affordance, abelha2017learning}.}. Thus, $m=2$ and the set of objects $O_a$ consists of two objects, i.e., $|O_a| = 2$. Further, the ordering of objects in $O_a$ indicates the correspondence of the objects to the action and grasp parts. 

For shape scoring, we seek to train models that can predict whether an input point cloud is suited for performing a specific action. We formulate this as a binary classification problem. We represent the shape of the input object point clouds using Ensemble of Shape Functions (ESF) \cite{wohlkinger2011ensemble} which is a 640-D vector, shown to perform well in representing object shapes for tools \cite{schoeler2015bootstrapping,nair2019autonomous}. We then train independent neural networks that take an input ESF feature, and output a binary label indicating whether the input shape feature is suited for performing a specific action. Thus, we train separate neural networks, one for each action\footnote{The advantage of the binary classification is that for new actions, additional networks can be trained independently without affecting other networks. More details on the training process is available at \cite{nair2019autonomous}}. More specifically, we train separate networks for the tools' action parts, e.g., the head of a hammer or the flat head of a spatula, and for a supporting function: ``\textit{Handle}'', which refers to the tools' grasp part, e.g., hammer handle. For the training, we used 3D tool models collected from online sources and ToolWeb \cite{abelha2017learning}. For additional details on the training process, we refer the reader to \cite{nair2019autonomous}.

For the score prediction, given a set of objects $O_a$ to be used for constructing the tool, let $\mathcal{K}$ denote the set of objects in $O_a$ that are candidates for the action parts of the final tool, and let $O_a - \mathcal{K}$ be the set of candidate grasp parts. Then the shape score $\phi^{s}_{shape}(O_a)$ is computed by using the trained networks as:
\begin{align}
\label{eqn:shape}
    \phi^{s}_{shape}(O_a) = \prod_{o_i \in \mathcal{K}}p(action|o_i) \prod_{o_i \in O_a-\mathcal{K}}p(handle|o_i)
\end{align}
Where, $p$ is the prediction confidence of the corresponding network. Thus, we combine prediction confidences for all action parts and grasp parts. For example, for the action ``join-hammer$(O_a)$'' where $O_a$ consists of two objects $(o_1, o_2)$, the shape score $\phi^{s}_{shape}(O_a) = p(hammer\_head|o_1)*p(handle|o_2)$. 

\subsubsection{Material Scoring}
\label{sec:material}

Material scoring seeks to predict the material fitness of the objects in $O_a$ for performing the action $a$. This is indicated by the $MaterialFit()$ function in Algorithm 2. In this work, we make three simplifying assumptions. Firstly, we consider the construction of rigid tools which covers a vast range of real-world examples \cite{myers2015affordance, abelha2016model}. Secondly, we consider the material properties of the action parts of the tool since the action parts are more critical to performing the action with the tool \cite{shrivatsav2019tool}. Lastly, we assume that the materials that are appropriate for different tools is provided a-priori, e.g., hammer heads are made of wood or metal (Shown in Table \ref{tab:materials}). Note that this information can also be obtained using common knowledge bases such as RoboCSE \cite{daruna2019robocse}. 

\setlength{\extrarowheight}{0.3em}
\begin{table}[t]
\centering
\begin{tabular}{|l|l|}
\hline
\textbf{Tool} & \textbf{Material (Action part)} \\ \hline
Hammer        & Metal, Wood                     \\ \hline
Screwdriver   & Plastic, Metal                  \\ \hline
Ladle         & Plastic, Wood, Metal            \\ \hline
Spatula       & Plastic, Wood, Metal            \\ \hline
Rake          & Plastic, Wood, Metal            \\ \hline
Squeegee      & Foam                            \\ \hline
\end{tabular}
\vspace{1em}
\caption{Table indicating appropriate materials for action parts of different tools}
\label{tab:materials}
\end{table}

For material scoring, we seek to train models that can predict whether an input material is suited for performing a specific action. We formulate this as a multi-classification problem. We represent the material properties of the object using spectral readings, since it has been shown to work well for material classification problems in prior work \cite{erickson2019classification, erickson2020multimodal,shrivatsav2019tool}. For extracting the spectral readings, the robot uses a commercially available handheld spectrometer\footnote{https://www.consumerphysics.com/}, called SCiO, to measure the reflected intensities of different wavelengths, in order to profile and classify object materials. The spectrometer generates a 331-D real-valued vector of spectral intensities. Then given a dataset of SCiO measurements from an assortment of objects, we train a neural network through supervised learning to output a class label indicating the material of the object (Details regarding the training procedure is available in \cite{erickson2019classification}).

For the material score prediction, given the spectral readings for the action parts in $O_a$ denoted by $\mathcal{K}$, we map the predicted class label to values in Table \ref{tab:materials} to compute the material score using the prediction confidence of the model. Let $T(a)$ denote the set of appropriate materials for performing an action $a$. Then the material score is computed as:
\begin{align}
\label{eqn:mat}
\phi^{s}_{mat}(O_a) = 
    \begin{cases}
        z = \displaystyle \prod_{o_i \in \mathcal{K}} \ \max_{c_i \in T(a)} p(c_i | o_i), & \text{if} \ z \geq \text{t} \\
        -\infty, & \text{otherwise}
    \end{cases}
\end{align}
Where, $p$ is the prediction confidence of the network regarding the class $c_i$. We compute the max prediction confidence across all the appropriate classes $c_i \in T(a)$, and their product over the action parts in $\mathcal{K}$. For example, for the action ``join-hammer$(O_a)$'', where $O_a$ consists of two objects $(o_1, o_2)$, the material score $\phi^{s}_{mat}(O_a) = \max(p(metal | o_1), p(wood | o_1))$. If the max value exceeds some threshold\footnote{We empirically determined a threshold of 0.6 to work well} denoted by $t$, then the corresponding value is returned. Otherwise, the model returns $-\infty$. Hence, note that material prediction acts as a hard constraint, by directly eliminating any objects that are made of inappropriate materials, thus reducing the potential search effort.

\subsubsection{Attachment Prediction}
\label{sec:attachment}
Given a set of objects, we seek to predict whether the objects can be attached to construct a tool. This is indicated by the $canAttach()$ function in Algorithm 2. In order to attach the objects, we consider three attachment types, namely, \textit{pierce attachment} (piercing one object with another, e.g., foam pierced with a screwdriver), \textit{grasp attachment} (grasping one object with another, e.g., a coin grasped with pliers), and \textit{magnetic attachment} (attaching objects via magnets on them). For magnetic attachments, we manually specify whether magnets are present on the objects, enabling them to be attached. For pierce and grasp attachment, we check whether the attachments are possible as described below. If no attachments are possible for the given set of objects, the feature score returns $-\infty$, indicating that the objects are not a viable combination. Thus, the search eliminates any objects that cannot be attached. 

\noindent \textbf{Pierce attachment:} Similar to material reasoning, we use the SCiO sensor to reason about material pierceability. We assume homogeneity of materials, i.e., if an object is pierceable, it is uniformly pierceable throughout the object. We train a neural network to output a binary label indicating pierceability of the input spectral reading \cite{nair2019autonomous}. If the model outputs 0, the objects cannot be attached via piercing. 

\noindent \textbf{Grasp attachment:} To predict grasp attachment, we model the grasping tool (pliers or tongs) as an extended robot gripper. This allows the use of existing robot grasp sampling approaches \cite{ten2017grasp, levine2018learning, zech2016grasp}, for computing locations where a given object can be grasped. In particular, we use the approach discussed by ten Pas et al., that outputs a set of grasp locations given the input parameters reflecting the attributes of the pliers or tongs used for grasping  \cite{ten2017grasp}. If the approach could not sample any grasp locations, the objects cannot be attached via grasping. 

\subsection{Incorporating the Sensor Trust Parameter}
In this section, we describe how the sensor trust parameter (Line 4, Algorithm 2) is incorporated to compute the feature score in two different ways. The first approach includes trusting the shape, material, and attachment predictions of the models described above. The second approach allows the robot to deal with possible false negatives in material and attachment predictions, by only incorporating the shape score into the feature score computation. 

\subsubsection{Fully trust sensors}
\label{sec:trust}
In the case that the robot fully trusts the material and attachment predictions, the trust parameter is set to $true$ (Line 4, Algorithm 2). The final feature score is then computed as a weighted sum of the shape and material scores, if the objects can be attached  (Algorithm 2, Lines 5-8). We found uniform weights of $\lambda_1 = 1, \lambda_2 = 1$, to work well for tool constructions. If the objects cannot be attached, then $\phi(s) = -\infty$, indicating that the objects in $O_a$ do not form a valid combination. Otherwise, using equations \ref{eqn:shape} and \ref{eqn:mat}:
\begin{align}
\label{eqn:score}
score(s, O_a) = \lambda_1 * \phi^{s}_{shape}(O_a) + \lambda_2 * \phi^{s}_{mat}(O_a)
\end{align}
Since both material and attachment predictions are hard constraints, certain object combinations can be assigned a score of $-\infty$, indicating that the robot does not attempt these constructions. As described before, this can lead to cases of false negatives where the robot is unable to find the correct construction due to incorrect material or attachment predictions. In these cases, the algorithm tracks the rejected object combinations in $O_{reject}$ (Algorithm 1, Line 26), and repeats the search as described below, by setting trust to $false$ (Algorithm 1, Lines 33-34).

\subsubsection{Not fully trust sensors}
\label{sec:notrust}

In case of false negatives, the robot can choose to eliminate the hard constraints of material and attachment prediction from the feature score computation, thus allowing the robot to explore the initially rejected object combinations by using only the shape score. This is achieved by setting the trust flag to $false$ in our implementation (Lines 14-15, Algorithm 2). In this case, we attempt to re-plan using the feature score as:
\begin{align}
\label{eqn:no_trust}
    \phi(s) =
    \begin{cases}
        \phi^{s}_{shape}(O_a), & \text{if} \ O_a \subseteq O_{reject}\\
        -\infty, & \text{otherwise}
    \end{cases}
\end{align}
Here, $O_{reject}$ indicates the set of objects that were initially rejected by the material and/or the attachment predictions. Since, shape score is a soft constraint, i.e., it does not eliminate any object combinations completely, we use the shape score to guide the search in case of the rejected objects. In the worst case, this causes the robot to explore all $^nP_m$ permutations of objects. However, as shown in our results, shape score can serve as a useful guide for improving tool construction performance in practice, when compared to naively exploring all possible object combinations. The final feature score computation, influenced by attachments and the trust parameter, can be summarized as follows from equations \ref{eqn:score}, \ref{eqn:no_trust}:
\begin{align*} 
\phi(s) = 
    \begin{cases}
        score(s, O_a), & \text{if attachable \& trust} \\
        \phi^{s}_{shape}(O_a), & \text{if \textbf{not} trust \& $O_a \subseteq O_{reject}$} \\
        -\infty, & \text{otherwise}
    \end{cases}
\end{align*}

\subsection{Final cost computation}
\label{sec:final}
Once the feature score is computed, the final cost function is computed as $f(s) = g(s) + h(s) - \lambda*\phi(s)$. Interestingly, we found that $\lambda=1$, thus $f(s) = g(s) + h(s) - \phi(s)$, performs very well with the choice of search algorithms and heuristics in this work for the problem of tool construction. In this case, the higher the feature score $\phi(s)$, the lower the cost $f(s)$, in turn guiding the search to choose nodes with higher feature score (lower $f(s)$ values). Additionally, the values of the feature score are within the range $0 \leq \phi(s) \leq 2$. Since we use existing planning heuristics that have been shown to work well, and the task plans generated have $\gg 2$ steps involved, $g(s) + h(s) \gg 2$ and thus, $f(s) > 0$. Thus, $\lambda=1$ works well for the problems described in this work. However, this presents an interesting research question for our future work in terms of an in-depth analysis of the choice of heuristic and feature score values, and its influence on the guarantees of the search.


\subsection{Implementation Details}
In this section, we describe additional details regarding the implementation of the work, both in terms of the algorithm, as well as the physical implementation on the robot.

\subsubsection{Algorithm Implementation}
In terms of implementation, the process begins with the trust parameter set to $True$. FGS generates a task plan that involves combining objects to construct the required tool. Once a task plan is successfully found, the robot can proceed with executing the task plan and joining the parts indicated by $O_a$ as described in \cite{nair2019autonomous}, to construct the required tool for completing the task. If the tool could not be successfully constructed or used, the plan execution is said to have failed, and the robot re-plans to generate a new task plan with a different object combination, since the algorithm tracks the attempted object combinations. Note that the approach also keeps track of object combinations rejected by material and attachment predictions in $O_{reject}$. If no solution could be found with trust set to $True$, and $O_{reject} \neq \emptyset$, then the robot switches trust to $false$, and FGS explores the object combinations within $O_{reject}$ (Lines 33-34 of Algorithm 1). If no solution could be found with either trust setting, FGS returns a complete failure and terminates.

Further note that in this work, we do not explicitly deal with symbol grounding \cite{harnad1990symbol} and symbol anchoring \cite{coradeschi2003introduction} problems. We overcome these issues by manually mapping the object point clouds to their specific symbols within the planning domain definition. Once the task plan is generated, the mapping is then used to match the symbols within the task plan to their corresponding objects in the physical world, via their point clouds. However, existing approaches can potentially be adapted in order to refine the symbol grounding functions \cite{hiraoka2018refining}, or to enable the robot to automatically extract the relationships between the object point clouds and their abstract symbolic representations \cite{konidaris2018skills}.

\subsubsection{Physical Implementation}
The spectrometer used in this work can be activated either using a physical button located on the sensor, or through an app that is provided with the sensor. However, pressing the physical button requires precision and careful application of the correct amount of force, which can be challenging for the robot since it may potentially damage the sensor if the applied force exceeds a certain threshold. To prevent this, in our implementation, the robot simply moves the scanner over the objects, and the user then manually presses a key within the app to activate the sensor. Additionally, the rate of scanning is also limited by the speed of the robot arm itself. Since the robot arm used in this work moves rather slowly, it took about $\approx$1.7 mins on average to scan 10 objects, while this would take less than 30 seconds for a human. Overcoming these issues and several other manipulation challenges are essential to ensure practical applicability of this work. We discuss this in more detail in Section \ref{sec:conclusion}.

\section{Experimental Validation and Results}
In this section, we describe our experimental setup and present our results alongside each evaluation. We validate our approach on three diverse types of tasks involving tool construction in a household domain, namely, wood-working, cooking, and cleaning. For wood-working tasks, the tools to be constructed include hammer and screwdriver; for cooking tasks the tools include spatula and ladle; and lastly for cleaning tasks the tools include rake and squeegee. Each tool is constructed from two parts ($m=2$) corresponding to the action and grasp parts of the tool. Our experiments seek to validate the following three aspects of our work:

\begin{enumerate}
    \item \textbf{Performance of feature guided $A^*$ against baselines:} In order to investigate the informativeness of including feature score in heuristic search, we evaluate the feature guided $A^*$ approach against three baselines. We also evaluate our approach in terms of the two different settings of the sensor trust parameter to investigate the benefits of introducing sensor trust.
    \item \textbf{Combining feature scoring with other heuristic search algorithms:} To investigate whether feature scoring can generalize to other search approaches, we integrate feature scoring with two additional heuristic search algorithms. Specifically, we present results combining feature scoring with weighted $A^*$ and enforced hill-climbing with the fast-forward heuristic \cite{hoffmann2001ff}.
    \item \textbf{Adaptability of task plans to objects in the robot's environment:} We evaluate whether the robot can adapt its task plans to appropriately use the constructed tool, as the objects available for construction are modified. This measures whether the robot can flexibly generate task plans in response to the objects in the environment.
\end{enumerate}

\begin{figure}[t]
	\centering
	\includegraphics[width=0.6\textwidth]{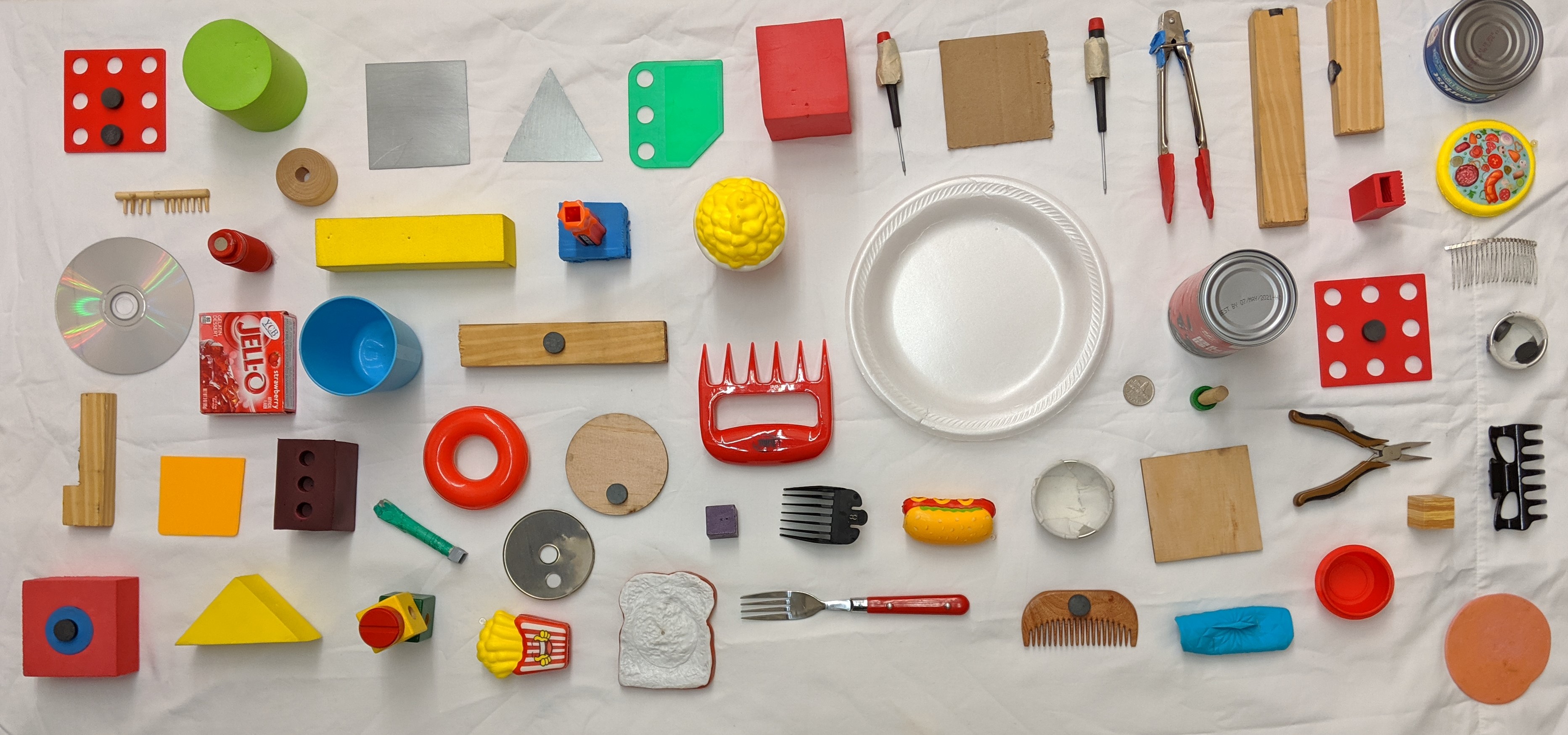}
	\caption{\normalsize Dataset of 58 objects used for the experiments, made of different materials}
	\label{fig:full_DS}
\end{figure}

For all our experiments, we use a test set consisting of 58 previously unseen candidate objects for tool construction (shown in Figure \ref{fig:full_DS}). These objects consist of metal (11/58), wood (12/58), plastic (19/58), paper (2/58) and foam (14/58) objects. Only the foam and paper objects are pierceable. Prior to planning, the robot scans the materials of the objects for material scoring and attachment predictions. For our results, we evaluate the statistical significance where it is applicable, using repeated measures ANOVA with post-hoc Tukey's test. We discuss each experiment in more detail below, along with the results for each. 

\subsection{Performance of Feature Guided A*}
\label{sec:expt1_1}

In this section, we evaluate the performance of feature guided $A^*$ against three baselines: i) standard $A^*$, where $f(s) = g(s) + h(s)$, ii) feature guided uniform cost search, where $f(s) = g(s) + 2.0 - \phi(s)$, and iii) standard uniform cost search, where $f(s) = g(s)$. In ii), we use $2.0 - \phi(s)$ to add a positive value to $g(s)$ since, $0 \leq \phi(s) \leq 2$. As heuristic with $A^*$, we use the cost optimal landmark heuristic \cite{karpas2009cost}. We also vary the sensor trust parameter, and present results for the two cases where the robot is not allowed to change the trust parameter (trust always set to $true$, i.e., lines 33-34 of Algorithm 1 not executed), and for the case where the robot can change it to $false$ when no plan is found.

For the evaluation, we create six different tasks, two tasks each for wood-working, cooking and cleaning. Each task requires the construction of \textit{one} specific tool for its completion, e.g., one of the tasks in wood-working requires construction of a hammer, and the other requires construction of a screwdriver. For each task we created 10 test cases, where each test case consisted of 10 objects chosen from the 58 in Figure \ref{fig:full_DS}, that could potentially be combined to construct the required tool. We report the average results across the test cases for each task type (total $10 \times 2$ cases per task type with 10 candidate objects per case). We create each test set by choosing a random set of objects, ensuring that only one ``correct'' combination of objects exists per set. The correct combinations are determined based on human assessment of the objects. For each task, we instantiate the corresponding domain and problem definitions in PDDL\footnote{In the planning problem definition, the objects are instantiated numerically through ``obj0'' to ``obj9'', where each literal is automatically grounded to one of the 10 objects during planning time. Our planning and domain definitions are available at \url{https://github.com/Lnair1993/Tool_Macgyvering}.}. 

The metrics used in this experiment include i) the \textit{number of nodes expanded} during search as a measure of computational resources consumed, ii) the \textit{number of failed construction attempts} before a working tool was found (also referred to as ``attempts'' in this paper), and iii) the \textit{success rate} indicating the number of times the robot successfully found a working tool. Ideally, we would like the number of nodes expanded and the number of failed construction attempts to be as low as possible. Note that the brute force number of failed construction attempts for 10 objects is 89, since there are $^{10}P_2$ possible object permutations for $m=2$, with 89 incorrect possibilities. Ideally, we would like the number of failed construction attempts to be 0. The success rate should be as high as possible, ideally equal to 100\%.

\setlength{\extrarowheight}{0.3em}
\begin{table*}[t]
\centering
\small
\begin{tabular}{c|c|c|c|c|c|c|c|c|c|c|c|c|}
\cline{2-13}
                                           & \multicolumn{4}{c|}{\textbf{Cleaning}}      & \multicolumn{4}{c|}{\textbf{Cooking}} & \multicolumn{4}{c|}{\textbf{Wood-working}}             \\ \cline{2-13} 
                                           & \textbf{FS+H} & H             & FS   & UCS  & \textbf{FS+H}  & H    & FS    & UCS   & \textbf{FS+H} & H             & FS         & UCS   \\ \hline
\multicolumn{1}{|c|}{\textbf{\# Nodes}}    & \textbf{5187} & \textbf{5187} & 9061 & 9061 & \textbf{329}   & 604  & 36237 & 36213 & 7264          & \textbf{6936} & 28606      & 28734 \\ \hline
\multicolumn{1}{|c|}{\begin{tabular}[c]{@{}l@{}}\textbf{\# Failed}\\ \textbf{Attempts}\end{tabular}} & \textbf{2}    & 46            & 3    & 49   & \textbf{3}     & 48   & 4     & 40    & \textbf{2}    & 45            & \textbf{2} & 37    \\ \hline
\end{tabular}
\caption{Table comparing feature guided $A^*$ (``FS+H'') with baselines. The other notations: ``H'' - standard $A^*$; ``FS'' - feature guided uniform cost search; ``UCS'' - standard uniform cost search. This table reports the \textit{average} number of failed attempts per task, across test cases where tool construction was successful. Note that the max number of failed attempts possible is 89 (brute force).}
\label{fig:expt1_tab1}
\end{table*}

\begin{figure}[t]
	\centering
	\includegraphics[width=1.0\textwidth]{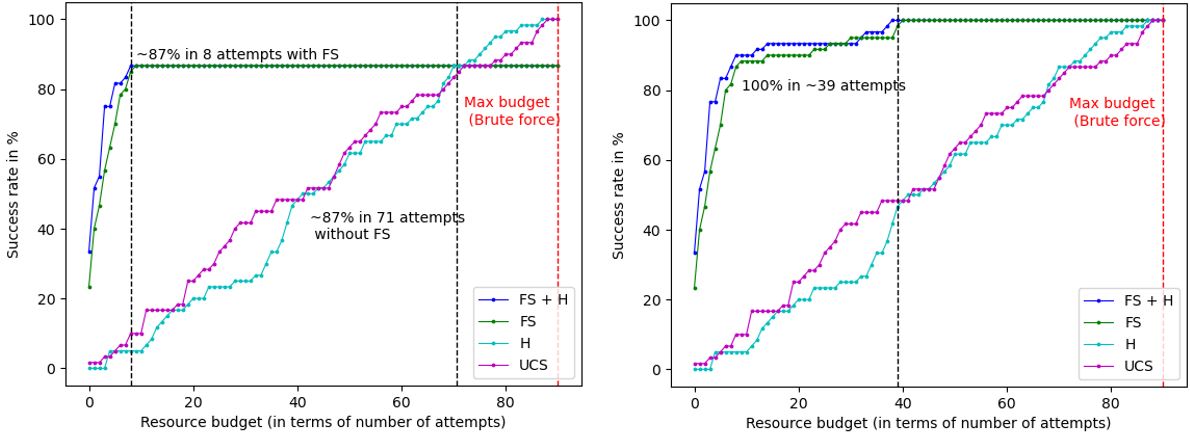}
	\caption{\normalsize Graphs highlighting the success rates for the two different modes of feature scoring based on sensor trust parameter, in relation to the number of failed attempts. For \textbf{left}, the sensors are fully trusted, and for the right, sensors are not fully trusted. Note that X-axis highlights the \textit{actual} number of attempts across all test cases for wood-working, cooking and cleaning put together.}
	\label{fig:graph}
\end{figure}


Table \ref{fig:expt1_tab1} shows the performance of feature guided $A^*$ (where $f(s) = g(s) + h(s) - \phi(s)$, denoted by ``FS+H'') compared to the different baselines: ``H'' denotes standard $A^*$ (where $f(s) = g(s) + h(s)$), ``FS'' denotes feature guided uniform cost search (where $f(s) = g(s) + 2.0 - \phi(s)$), and ``UCS'' denotes standard uniform cost search (where $f(s) = g(s)$). The values reported per task are the average performances across the test cases where tool constructions were successful. As shown in Table \ref{fig:expt1_tab1}, incorporating feature scoring (FS, FS+H) helps significantly reduce the number of failed construction attempts compared to the baselines without feature scoring (H, UCS), with $p < 0.001$. Since heuristics can help reduce the search effort in terms of number of nodes expanded, we see that approaches that do not use heuristics (FS and UCS) expand significantly more nodes than FS+H and H, with $p < 0.001$. Note that there is no statistically significant difference in the number of nodes expanded between H and FS+H. Thus, using feature scoring with heuristics (FS+H) yields the best performance in terms of \textit{both} number of nodes expanded, and the number of failed construction attempts. \textbf{To summarize, these results show that feature scoring is informative to heuristic search by significantly reducing the average number of failed construction attempts to $\approx$ 2 compared to $\approx$ 46 without it (brute force number of failed attempts is 89).} 

Further, in figure \ref{fig:graph}, we plot the success rate vs. the resource budget of the robot in terms of the \textit{permissible} number of failed attempts. That is, the robot is not allowed to try any more than a fixed number of attempts, indicated by the resource budget. Figure \ref{fig:graph}:left considers the case where the sensor trust parameter is always set to $true$, and Figure \ref{fig:graph}:right considers the case when the robot is allowed to switch the trust to $false$, if a solution was not found. Note that in contrast to Table \ref{fig:expt1_tab1}, the graphs report \textit{actual} number of failed attempts, \textit{across all tasks}, whereas Table \ref{fig:expt1_tab1} reports the \textit{average} number of failed attempts across the test cases per task, for tool constructions that were successful. In Figure \ref{fig:graph}:left, we see that FGS (FS+H and FS) achieves a success rate of 86.67\% (52/60 constructions) within a resource budget of $\approx$ 8 failed attempts to do so. This indicates that 13.33\% of the valid constructions were treated as false negatives by material and attachment predictions, and were completely removed from consideration (unattempted). Thus, increasing the permissible resource budget beyond 8, does not make any difference. Without feature scoring, H and UCS achieve a success rate of 87\% with a budget of 71 attempts, and 100\% after exploring nearly every possible construction (max resource budget of 89 failed attempts). In contrast, when the robot is allowed to switch the trust parameter, the robot uses shape scoring alone to continue guiding the search. As shown in Figure \ref{fig:graph}:right, FGS (FS+H and FS) achieves 100\% success rate within a budget of $\approx$ 39 attempts, since the robot does not eliminate any object combinations from consideration. The performance is also significantly better than the baselines that do not use feature scoring. This is because shape scoring guides the search through the space of object combinations based on the objects' shape fitness, compared to H and UCS that do not have any measure of the fitness of the objects for tool construction. \textbf{To summarize, feature scoring enables the robot to successfully construct tools by leveraging the sensor trust parameter, while significantly outperforming the baselines in terms of the resource budget required.}

\setlength{\extrarowheight}{0.3em}
\renewcommand{\tabcolsep}{3.5pt}
\begin{table}[t]
\centering
\small
\begin{tabular}{c|c|c?c|c?c|c?c|c?c|c?c|c|}
\cline{2-13}
                                                                                             & \multicolumn{4}{c?}{\textbf{Cleaning}}                                         & \multicolumn{4}{c?}{\textbf{Wood-working}}                                            & \multicolumn{4}{c|}{\textbf{Cooking}}                                         \\ \cline{2-13} 
                                                                                             & \multicolumn{2}{c?}{\textbf{Squeegee}} & \multicolumn{2}{c?}{\textbf{Rake}}    & \multicolumn{2}{c?}{\textbf{Hammer}}  & \multicolumn{2}{c?}{\textbf{Screwdriver}} & \multicolumn{2}{c?}{\textbf{Spatula}} & \multicolumn{2}{c|}{\textbf{Ladle}}   \\ \cline{2-13} 
                                                                                             & \textbf{trust}  & \textbf{$\sim$trust} & \textbf{trust} & \textbf{$\sim$trust} & \textbf{trust} & \textbf{$\sim$trust} & \textbf{trust}   & \textbf{$\sim$trust}   & \textbf{trust} & \textbf{$\sim$trust} & \textbf{trust} & \textbf{$\sim$trust} \\ \hline
\multicolumn{1}{|c|}{\textbf{\begin{tabular}[c]{@{}c@{}}\# Failed \\ Attempts\end{tabular}}} & 0               & 1                    & 3              & 7                    & 2              & 2                    & 2                & 8                      & 3              & 7                    & 2              & 2                    \\ \hline
\multicolumn{1}{|c|}{\textbf{\# Success}}                                                    & 9/10            & 10/10                & 7/10           & 10/10                & 10/10          & 10/10                & 8/10             & 10/10                  & 8/10           & 10/10                & 10/10          & 10/10                \\ \hline
\end{tabular}
\vspace{1em}
\caption{Table showing tool-wise breakdown in performance for feature guided $A^*$. This table reports the \textit{average} number of failed attempts per tool, across cases where tool construction was successful. The notation $\sim$trust indicates cases where sensors are \textit{not} fully trusted. Note that max \# failed attempts is 89.}
\label{fig:expt1_tab3}
\end{table}

\begin{figure}[t]
	\centering
	\includegraphics[width=0.95\textwidth]{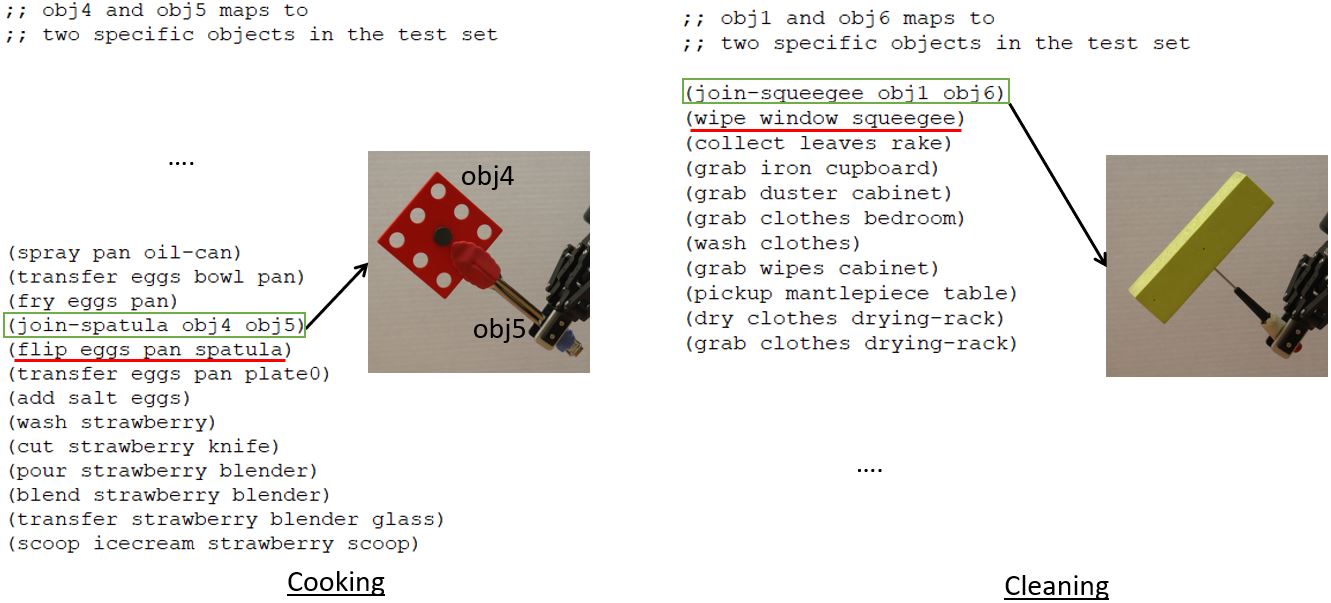}
	\caption{\normalsize (Left) A sample task plan where a spatula must be constructed for a cooking task, and the planner uses the flat piece (obj4 in the problem definition), and tongs (obj5 in the problem definition). The action ``join-spatula'' refers to the construction of the spatula using obj4 and obj5. Similarly, (right) a squeegee is constructed from obj1 (foam block) and obj6 (screwdriver) for the cleaning task. Without tool construction (in green) the actions underlined in red would fail.}
	\label{fig:expt1_plan}
\end{figure}

In order to understand which tools were more challenging for feature scoring, Table \ref{fig:expt1_tab3} shows a tool-wise breakdown in performance for feature guided $A^*$ for the two different sensor trust values. The notation ``trust'' denotes the case where sensors are fully trusted, and ``$\sim$trust'' denotes case where they are not fully trusted. When the sensors are fully trusted, rakes were a particularly challenging test case, as indicated by the lowest success rate of 7/10. In contrast, hammers and ladles have a success rate of 10/10. The failure cases for each tool arises from incorrect material and attachment predictions. While not fully trusting the sensors ($\sim$trust) leads to a 100\% success rate (60/60 cases), using shape score alone leads to more failed construction attempts when compared to combining shape with material and attachment predictions since shape alone is less informative (e.g., for rake, $\sim$trust has 7 failed attempts vs. 3 attempts for trust).

Figure \ref{fig:expt1_plan} shows sample task plans generated by the robot in cooking and cleaning tasks. In the case of cooking, the robot needed a spatula to flip the eggs, and used a flat piece (obj4) with tongs (obj5) to construct the spatula via grasp attachment. For cleaning, the robot needed a squeegee to clean the window, and used a foam block (obj1) and screwdriver (obj6) to construct the squeegee via pierce attachment. Without the constructed tools, the actions highlighted in red would fail, i.e., the ``flip'' action would fail without the constructed spatula. Hence, \textbf{FGS enables the robot to replace missing tools through construction. To summarize, the key findings of this experiment indicate that feature scoring is highly informative for heuristic search by reducing the number of nodes expanded by $\approx$ 82\%, and the number of failed construction attempts by $\approx$ 93\%, compared to the baselines. Further, allowing the robot to switch the trust parameter when a plan is not found, helps achieve a success rate of 100\% within a budget of $\approx$ 39 attempts, significantly outperforming baselines that do not use feature scoring.}

\subsection{Feature Scoring With Other Heuristic Search Algorithms}
\begin{table*}[]
\centering
\small
\begin{tabular}{c|c|c|c|c|c|c|c|c|c|}
\cline{2-10}
                                           & \multicolumn{3}{c|}{\textbf{Cleaning}}                   & \multicolumn{3}{c|}{\textbf{Cooking}}                    & \multicolumn{3}{c|}{\textbf{Wood-working}}                   \\ \cline{2-10} 
                                           & \textbf{$A^*$+LM} & \textbf{w$A^*$+FF} & \textbf{eHC+FF} & \textbf{$A^*$+LM} & \textbf{w$A^*$+FF} & \textbf{eHC+FF} & \textbf{$A^*$+LM} & \textbf{w$A^*$+FF} & \textbf{eHC+FF} \\ \hline
\multicolumn{1}{|c|}{\textbf{\# Nodes}}    & 5187             & 21                & 21                & 329              & 23                & 35                & 7264             & 25                & 38                \\ \hline
\multicolumn{1}{|c|}{\begin{tabular}[c]{@{}l@{}}\textbf{\# Failed}\\ \textbf{Attempts}\end{tabular}} & 2                & 2                 & 4                 & 3                & 3                 & 4                 & 2                & 1                 & 2                 \\ \hline
\multicolumn{1}{|c|}{\textbf{Plan length}} & 20               & 22                & 22                & 19               & 19                & 19                & 11               & 15                & 15                \\ \hline
\end{tabular}
\caption{Table showing performance of feature guided Weighted $A^*$ (w$A^*$) and feature guided Enforced Hill-Climbing (eHC) with the fast-forward heuristic (FF).}
\label{fig:expt2_tab1}
\end{table*}

To demonstrate that feature scoring generalizes to other search approaches, in this section we present results for combining feature scoring with weighted $A^*$ \cite{pohl1970heuristic}, and enforced hill-climbing using the fast-forward heuristic \cite{hoffmann2001ff}. We use the same experimental setup and metrics as described in Section \ref{sec:expt1_1}. In addition, we also measure the output plan length to investigate the optimality of the different approaches. For weighted $A^*$, feature scoring is incorporated as $f(s) = g(s) + w*(h(s) - \phi(s))$, where $w$ indicates a weight parameter\footnote{Weight was set to 5.0}. For enforced hill-climbing, the cost function is computed as $f(s) = h(s) - \phi(s)$. For both weighted $A^*$ and enforced hill-climbing, we use the fast-forward heuristic, which has been shown to be successful for planning tasks in prior work \cite{hoffmann2001ff}.

In Table \ref{fig:expt2_tab1}, we present the results for feature scoring combined with $A^*$ and the cost-optimal landmark heuristic (``$A^*$+LM''), weighted $A^*$ with fast-forward heuristic (``w$A^*$+FF''), and enforced hill-climbing with fast forward heuristic (``eHC+FF''). Compared to $A^*$+LM, w$A^*$+FF and eHC+FF reduce the computational effort (fewer nodes expanded) in return for sub-optimal solutions (longer plan lengths). This is expected of weighted $A^*$ and enforced hill-climbing since they are inadmissible algorithms. There is no statistically significant difference between \# failed construction attempts in each case. \textbf{To summarize, the key finding of this experiment is that feature scoring can be applied to other planning heuristics such as fast-forward, and other heuristic search algorithms like weighted $A^*$ and enforced hill-climbing, to further reduce computational effort, albeit at the cost of optimality in terms of plan length.} 

\subsection{Adaptability of Task Plans}
In this section we evaluate the adaptability of our FGS approach to generate task plans based on objects in the environment, to appropriately use the constructed tool. We create three tasks, one task each for wood-working, cooking and cleaning. In each of the tasks, \textit{either} of two tools can be constructed to successfully complete the task, but there is only one ground truth depending on the objects available for construction. That is, the available objects only enable the construction of one of the two tools. Thus, the robot has to \textit{correctly choose the tool to be constructed}. In addition, the robot must adapt the task plan to appropriately use the constructed tool. For the wood-working task either a hammer (with action ``hit'') or a screwdriver (with action ``tighten'') can be used to attach two pieces of wood; for the cooking task either a spatula (with action ``flip'') or a ladle (with action ``scoop'') can be used to flip eggs; and for the cleaning task, either a squeegee (with action ``reach'') or a rake (with action ``collect'') can be used to collect garbage. 

\begin{figure}[t]
	\centering
	\includegraphics[width=0.6\textwidth]{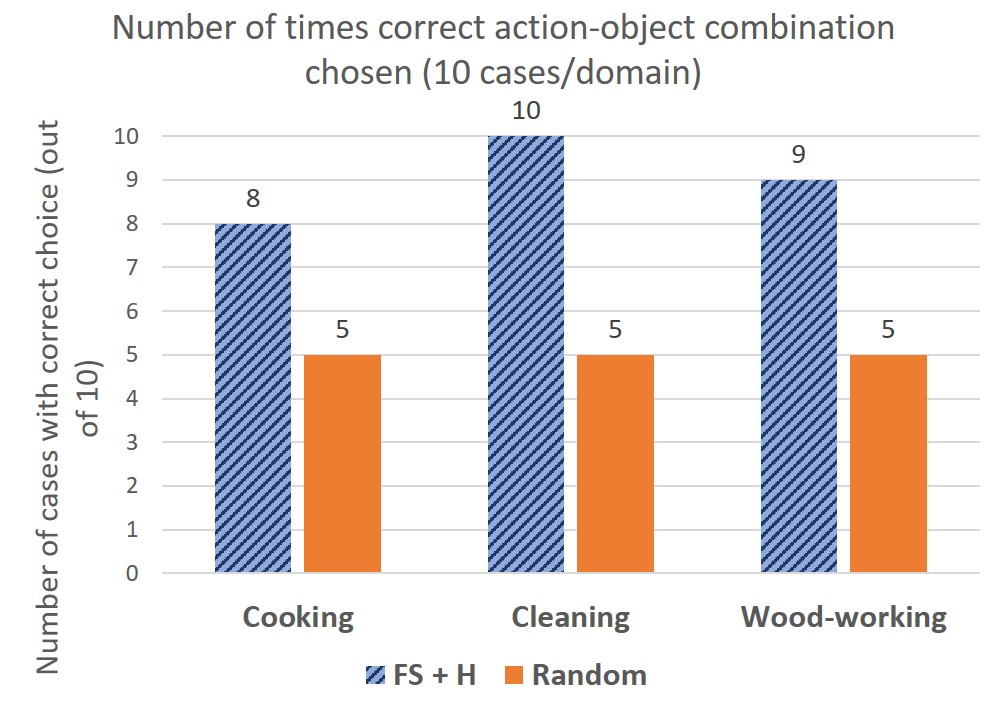}
	\caption{\normalsize Graph highlighting the number of times the correct object combination was chosen, compared to the random selection baseline. FGS outperforms random baseline with a statistical significance of $p < 0.01$.}
	\label{fig:expt3_graph1}
\end{figure}

For the evaluation, we create three different tasks, one each in wood-working, cooking, and cleaning. For each task, either one of two tools can be used to complete the task as described above. For each task, we created 10 different test sets of random objects, similar to the experiment described in Section \ref{sec:expt1_1}. In each case, only one ``correct'' combination exists. Thus, the robot has to correctly identify which of the two tools can be constructed for accomplishing the task, given the set of objects. We evaluate the performance of feature guided $A^*$ in each case alongside a random selection baseline to demonstrate the difficulty of the problem. The random selection baseline randomly chooses one of the two tool construction options for each task. Note that for each task, the domain and problem definitions are unchanged across the 10 test cases of objects. This indicates that the task plan adaptability does not require any manual modifications by the user, instead is the direct result of the sensor inputs received by the robot. 

The key metric used in this experiment includes the number of times the robot chose the correct tool to construct for each task. Thus, if the robot chose to construct a hammer, when the correct combination was to construct a screwdriver, the attempt is considered to have failed. We also present qualitative results showing some of the sample task plans and tools constructed by the robot for different sets of objects. 

Figure \ref{fig:expt3_graph1} shows the performance of feature guided $A^*$ compared to the random selection baseline. We see that feature guided $A^*$ chooses the correct tool for 27/30 cases, and significantly outperforms the random selection baseline ($p < 0.01$). The failure cases in the wood-working task arise due to noisy material detection. In the case of cooking task, the noisy point clouds sensed by the RGBD camera leads to incorrect choices, e.g., the concavity of bowls was not correctly detected for some ladles. 

\begin{figure}[t]
	\centering
	\includegraphics[width=1.0\textwidth]{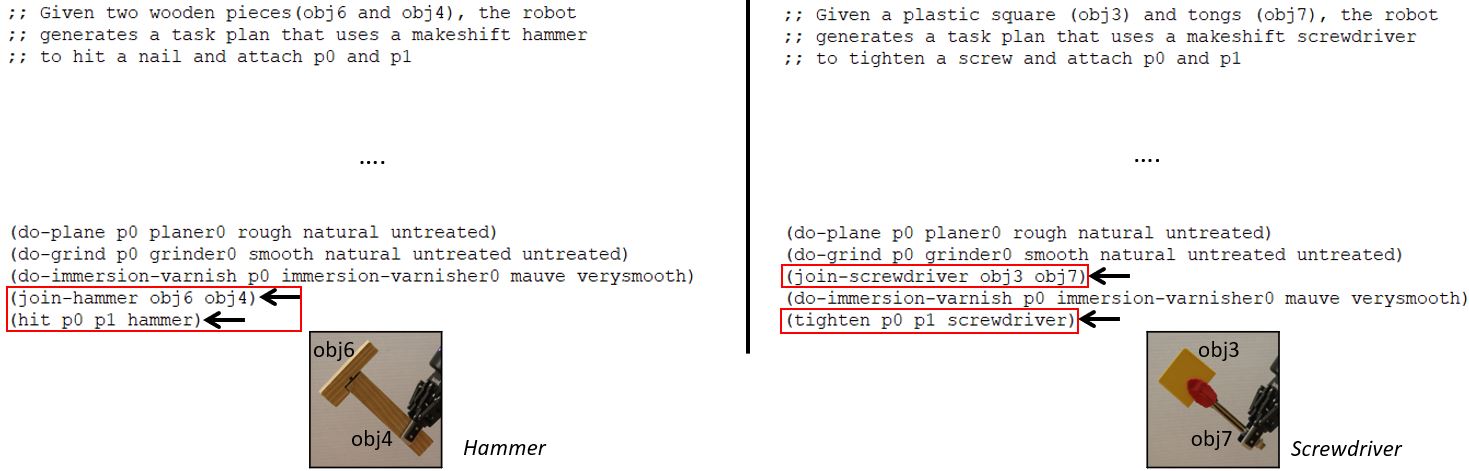}
	\caption{This figure shows the results for two of the test cases in wood-working. The task plans are adapted based on the constructed tool (i.e., hammer or screwdriver), to either ``hit'' or ``tighten'' to attach the two pieces of wood $p0$ and $p1$. Arrows denote the parts of the task plan that are adapted.}
	\label{fig:expt3_adaptPlan}
\end{figure}

\begin{figure}[t]
	\centering
	\includegraphics[width=0.95\textwidth]{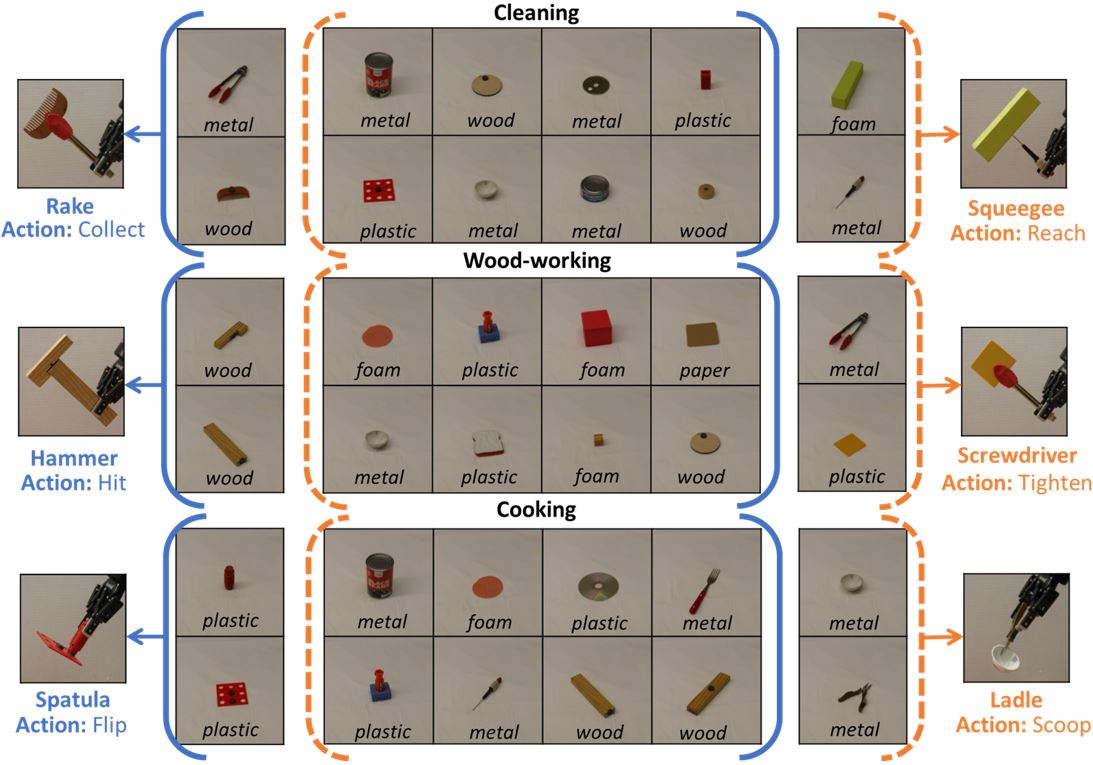}
	\caption{Collage indicating sample tool constructions output for two test cases per task. The solid and dashed brackets indicate the test set of objects provided in each case, along with the tool constructed for it. As the objects are changed, the corresponding constructed tool and action is different. Note that the problem and domain definitions are fixed for each task, and unchanged across the test cases per task.}
	\label{fig:expt3_collage}
\end{figure}

In Figure \ref{fig:expt3_adaptPlan}, we show two task plans that are generated within the task of wood-working. For the same task, either a hammer or a screwdriver can be used to attach two pieces of wood $p0$ and $p1$. Depending, on the objects available in the environment, the robot chooses to construct one of the two tools and adapts the task plan to use the corresponding tool for completing the task. As shown in the left of Figure \ref{fig:expt3_adaptPlan}, the robot chose to construct a hammer to ``hit'' and attach the two pieces of wood. Whereas, shown in the right of Figure \ref{fig:expt3_adaptPlan}, the robot chose to construct a screwdriver to ``tighten'' and attach the two pieces of wood. Similar adaptations are observed for the remaining two tasks as well: ``scoop'' with ladles vs. ``flip'' with spatulas in the cooking task, and ``reach'' with squeegees vs. ``collect'' with rakes in the cleaning task. Thus, the constructed tool depends on the objects in the environment, which in turn adapts the generated task plan to appropriately use the constructed replacement tool. 

In Figure \ref{fig:expt3_collage}, we present some qualitative results for six different tools constructed by the robot for six of the test cases. The solid and dashed parentheses highlight the input test set. For example, given the metal bowl and metal pliers, the robot chooses to construct a ladle (and use the ``scoop'' action in the task plan). In contrast, when the pliers and bowl are replaced with a plastic handle and a flat plastic piece, the robot chooses to construct a spatula instead (and use the ``flip'' action in the task plan). Given that the problem and domain definitions are unchanged for the two cases, this shows that the robot is able to adapt the task plan in response to the objects in the environment. \textbf{To summarize, the key finding of this experiment is that the robot is able to successfully adapt the task plan to construct and use the appropriate tool depending on the objects available for construction, with an accuracy of 90\% (27/30 cases).}

\section{Conclusion and Future Work}
\label{sec:conclusion}

In this work, we presented the Feature Guided Search (FGS) approach that allows existing heuristic search algorithms to be efficiently applied to the problem of tool construction in the context of task planning. Our approach enables the robot to effectively construct and use tools in cases where the required tools for performing the task are unavailable. We relaxed key assumptions of the prior work in terms of eliminating the need to specify an input action, instead integrating tool construction within a task planning framework. Our key findings can be summarized as follows:
\begin{itemize}
    \item FGS significantly reduces the number of nodes expanded by $\approx 82\%$, and the number of construction attempts by $\approx 93\%$, compared to standard heuristic search baselines. 
    \item The approach achieves a success rate of 87\% within a resource budget of 8 attempts when sensors are fully trusted, and 100\% within a budget of 39 attempts, when the sensors are not fully trusted.
    \item FGS enables flexible generation of task plans based on objects in the environment, by adapting the task plan to appropriately use the constructed tool.
    \item Feature scoring can also be effectively combined with other heuristic search algorithms such as weighted $A^*$ and enforced hill-climbing.
\end{itemize}

Our work is one of the first to integrate tool construction within a task planning framework, but there remain many unaddressed manipulation challenges in tool construction that are beyond the scope of this paper. Tool construction is a challenging manipulation problem that involves appropriately grasping and combining the objects to successfully construct the tool. That is, once the robot has correctly identified the objects that need to be combined (focus of this paper), the robot would then have to physically combine the objects, and use the constructed tool for the task. Currently, our work pre-specifies the trajectories to be followed for tool construction, although existing research in robot assembly can be leveraged to potentially accomplish this \cite{thomas2018learning}. Further, a key question to be addressed is, \textit{how can the robot learn to appropriately use the constructed tool?} Future work could address this problem by leveraging existing research in tool use \cite{stoytchev2005behavior, sinapov2008detecting, sinapov2007learning}, and trajectory-based skill adaptation \cite{fitzgerald2014representing, gajewski2019adapting}. Upon successful construction of the tool, the research problem reduces to that of using the tool appropriately. In this case, the robot can either learn how to use the tool as described in \cite{stoytchev2005behavior, sinapov2008detecting, sinapov2007learning} or, the robot can adapt previously known tool manipulation skills to the newly constructed tool as described in \cite{fitzgerald2014representing, gajewski2019adapting}. Addressing these challenges is important to further ensure practical applicability of tool construction.

Additionally, creation of tools through the attachment types discussed in this work is currently restricted to a limited number of use cases, in which two objects that have the specific attachment capabilities already exist, and are available to the robot. In the future, we seek to expand to more diverse types of attachments, including gluing or welding the objects together, as well as creation of tools from deformable materials, in order to improve the usability of our work. We further seek to expand on this work by investigating the application of feature scoring to domains other than tool construction. In particular, we seek to investigate the different ways in which feature score can be effectively combined with the cost function for other domains involving tool-use such as tool substitution. While our proposed cost function is dependent on the values of the feature score and is shown to perform well for tool construction, it is important to further investigate the cost function and its influence on the guarantees of the search to allow for a more generalized application of FGS.

FGS enables the robot to perform high-level decision making with respect to the objects that must be combined in order to construct a required tool. In this work, we use physical sensors (RGBD sensors and SCiO spectrometer) that produce partial point clouds and noisy spectral scans, leading to some challenges that commonly arise in the real world. Nevertheless, there are several open research questions that need to be addressed before this work can be deployed in a real setting. Thus, FGS is the first step within a larger pipeline, and we envision this work to be complementary to existing frameworks that are aimed at resilient and creative task execution, such as \cite{antunes2016human, stuckler2016mobile}. In summary, FGS presents a promising direction for dealing with tool-based problems in the area of creative problem solving.

\section{Acknowledgement}
The authors would like to thank Dr. Christopher G. Atkeson and Dr. Kalesha Bullard for their valuable feedback and insights on this work. 

\section{Conflict of Interest Statement}
The authors declare that the research was conducted in the absence of any commercial or financial relationships that could be construed as a potential conflict of interest.

\section{Funding}
This work is supported in part by NSF IIS 1564080 and ONR N000141612835.

\section{Data Availability Statement}
The source code and datasets developed for this work can be found at the Github page \url{https://github.com/Lnair1993/Tool_Macgyvering}.

\bibliographystyle{plain}  
\bibliography{references}  

\end{document}